\pdfoutput=1
\documentclass[11pt]{article}

\usepackage{acl}

\usepackage{times}
\usepackage{latexsym}
\usepackage{amsmath, amssymb}
\usepackage{dsfont}
\usepackage{tabularx}
\usepackage{graphicx}
\usepackage{booktabs}
\usepackage{makecell}
\usepackage{soul}
\usepackage{url}

\usepackage[T1]{fontenc}
\usepackage[utf8]{inputenc}

\usepackage{microtype}

\title{BinaryAlign: Word Alignment as Binary Sequence Labeling}

\author{Gaetan Lopez Latouche \and Marc-André Carbonneau \and Ben Swanson \\
        Ubisoft La Forge \\
        \{gaetan.lopez-latouche,marc-andre.carbonneau2,ben.swanson2\}@ubisoft.com}

\begin{document}
\maketitle
\begin{abstract}

Real world deployments of word alignment are almost certain to cover both high and low resource languages.  However, the state-of-the-art for this task recommends a different model class depending on the availability of gold alignment training data for a particular language pair.  We propose BinaryAlign, a novel word alignment technique based on binary sequence labeling that outperforms existing approaches in both scenarios, offering a unifying approach to the task.   Additionally, we vary the specific choice of multilingual foundation model, perform stratified error analysis over alignment error type, and explore the performance of BinaryAlign on non-English language pairs. We make our source code publicly available.\footnote{\href{https://github.com/ubisoft/ubisoft-laforge-BinaryAlignWordAlignementasBinarySequenceLabeling}{https://github.com/ubisoft/ubisoft-laforge-BinaryAlignWordAlignementasBinarySequenceLabeling}}

\end{abstract}

\section{Introduction}

% \begin{itemize}
%     \item define word alignment in one sentence 
%     \item tell what is it currently useful for: annotation projection, xml markups blabla.
% \end{itemize}

%Word alignment refers to the process of recognizing matching words in parallel texts.

% \bfs{1) Word alignment is a valid task
% 2) The are high and low resource languages
% 3) People use one type of model for full/few and one for zero
% 4) One method is better for one and another for another
% 5) but actually both are flawed
% 6) we have a new one that is better than all existing approaches in both domains
% My suggestions : number 3 and 4 seem to be the same point.  Add some indication that we also did extra experiments (this is important!!) Finally, Consider adding one short paragraph before everything that quickly lays out the series of points you're going to make (so that people can follow the logic at a high level as they read)}

Word alignment refers to the task of uncovering word correspondences between translated text pairs. The automatic prediction of word alignments dates back to the earliest work in machine translation with the IBM models \cite{brown1993mathematics} where they were used as hidden variables that permit the use of direct token to token translation probabilities.  While state of the art machine translation techniques have largely abandoned the use of word alignment as an explicit task \cite{li2022word} other use cases for alignments have emerged including lexical constraint incorporation \cite{chen2021lexically}, analysing and evaluating translation models \cite{bau2018identifying, neubig2019compare}, and cross-lingual language pre-training \cite{chi-etal-2021-improving}.

In many real-world applications word alignment must be performed across several languages, often including languages with manually annotated word alignment data and others lacking such annotations. We refer to those languages as high and low-resource languages respectively. While word alignment for high-resource languages can be learned in a few-shot or fully supervised setting depending on the amount of data, for low-resource languages zero-shot learning strategies must be employed due to data scarcity.

% In many real-world applications word alignment must be perform across many languages, oftentimes including low-resource languages. While word alignment for high-resource languages can be learned in a fully supervised manner from large quantities of annotated data or in a few, for low-resources languages few-shot and zero-shot learning strategies must be employed due to data scarcity.

\begin{figure}
    \centering
    \includegraphics[width=.4\textwidth ]{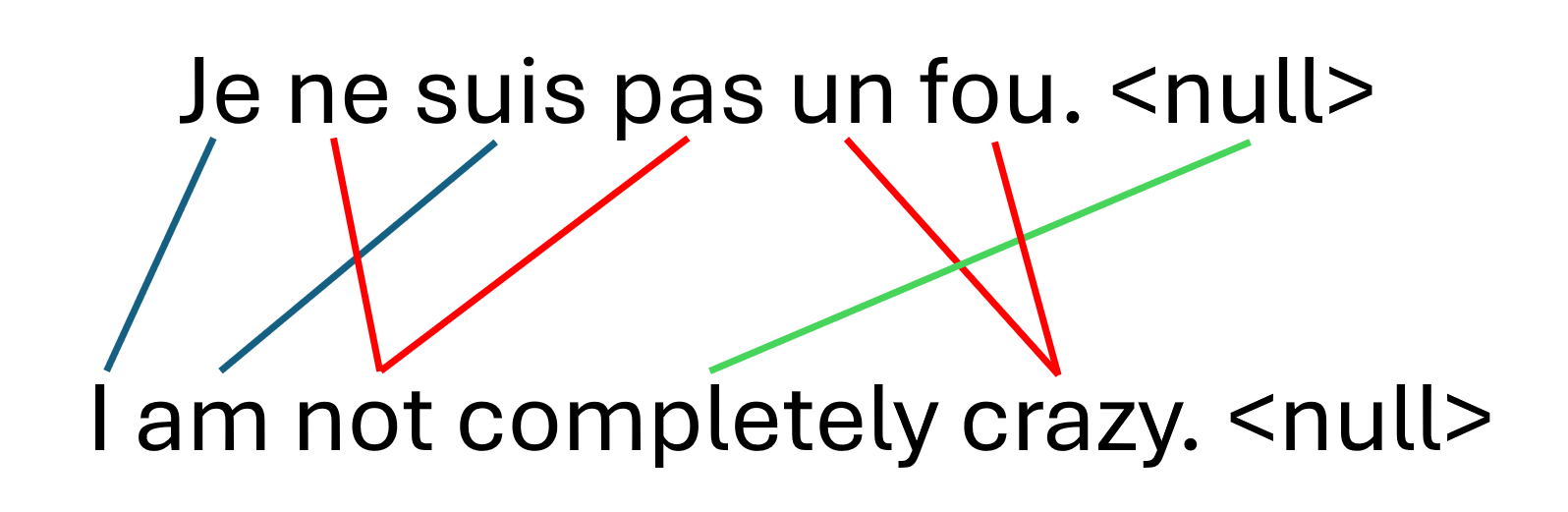}
    \caption{Example of alignment of an approximate translation, as often encountered in real-world applications. Links in red indicate situations where one word is aligned with several contiguous or non-contiguous words. The green line represent a situation where a word is untranslated which happens in many language pairs.}
    \label{fig:example}
\end{figure}

% CAPTION too big: When these words are not contiguous, as it often happens (e.g. French negations, approximate translations) span methods must predict multiple spans, while softmax methods must allows for multiple alignment per words. In both cases this relies on brittle hyper-parameters and threshold tuning. 

\begin{figure*}
    \centering
    \includegraphics[width=.9\textwidth]{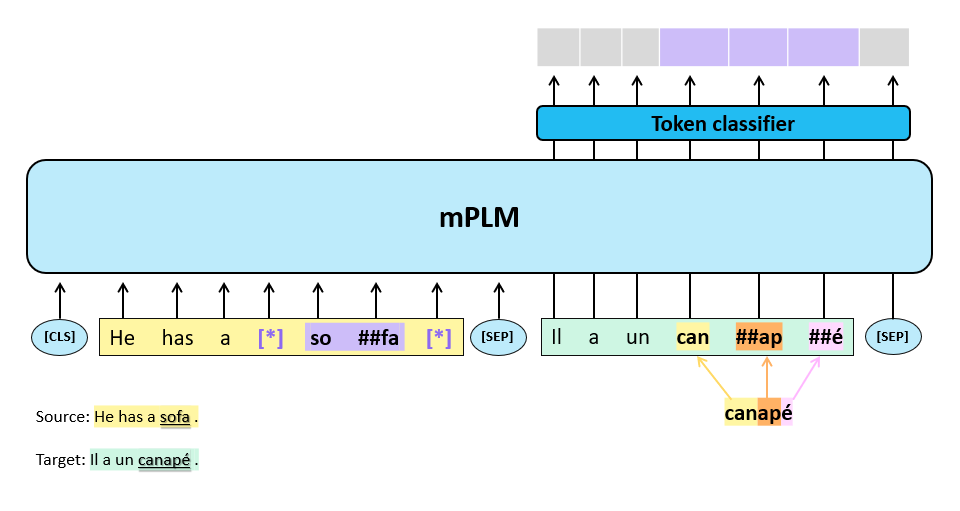}
    \caption{Illustration of our formalization of word alignment. In this example, the word "sofa" in the source sentence and the word "canapé" in the target sentence are aligned.}
    \label{fig:model}
\end{figure*} 

State-of-the-art supervised techniques formalize the task of word alignment as a collection of SQuAD-style span prediction problems \cite{nagata-etal-2020-supervised, wu-etal-2023-wspalign} while in zero-shot settings the best performing methods induce word alignment from the contextualized word embeddings of mulitingual pre-trained language models (mPLMs) \cite{jalili-sabet-etal-2020-simalign, dou-neubig-2021-word, wang-etal-2022-multilingual}. From a practical perspective, this discrepancy in the preferred method adds complexity to the deployment of word alignment models in real-world applications where both high and low-resource languages must be supported.

% However, most word alignment methods are suitable either in supervised or in zero-shot settings which adds complexity to the deployment of word alignment models in real-world applications where high and low-resource languages must be supported. To be specific, state-of-the-art supervised word alignment techniques formalize the task as a collection of SQuAD-style span prediction problems \cite{nagata-etal-2020-supervised, wu-etal-2023-wspalign} while in zero-shot settings, best performing methods induce word alignment from the contextualized word embeddings of mulitingual pre-trained language models (mPLMs) \cite{jalili-sabet-etal-2020-simalign, dou-neubig-2021-word, wang-etal-2022-multilingual}. 

% Most word alignment methods are suitable either for high-resource or low-resource languages. This complexifies the deployment of such models in real-world applications where many pairs of languages must be supported.

We observe a deeper issue that both span prediction and contextualized word embeddings are sub-optimal as each induces a bias in word alignment models that limits their accuracy. Span prediction methods cannot robustly deal with discontinuous word alignments without relying on complex post-processing and hyper-parameter tuning. Contextualized word embeddings method cannot deal effectively with untranslated words and one-to-multiple alignments because they rely on a softmax function that normalizes predictions at a sentence-level while in word alignment; one token being aligned to token $T$ does not mean that another token is less likely to be aligned to $T$. This poses word alignment as a single-label classification problem, while in reality it is better viewed as a series of binary classifications applied to each possible pair of words. Figure \ref{fig:example} shows some cases of one-to-multiple alignments, non-contiguous spans and untranslated words.

% cannot deal with untranslated words and poses word alignment as a single-label classification problem, while in reality, it is better viewed as a series of binary classifications applied to each possible pair of words. Figure \ref{fig:example} shows some cases of one-to-multiple alignments, non-contiguous spans and untranslated words.

%Moreover, those two formulations of word alignment are not designed for this task and have performance ceiling. Without complex post-processing, span prediction cannot deal with discontinuous word alignments. Also, contextualized word embeddings method rely on a softmax function that formalizes word alignment as a multi-class classification problem, which is different from the reality of the word alignment task where word X being aligned to word Y do not imply that word X is not aligned to word Z. 

%In this paper, we present a novel word alignment training and inference strategy that can be used in all real word settings. In particular, we reformulate the word alignment problem into a word based binary token classification, a new method designed for the word alignment task and that avoids the performance ceiling of previous works. By extensive experiments on five different language pairs, we show that our method outperforms previous works in all settings. 

In this paper, we present BinaryAlign, a novel word alignment solution that outperforms the state-of-the-art in zero-shot, few-shot and fully-supervised settings. In particular, we reformulate word alignment as a set of binary classification tasks in which an individual alignment prediction is made for each possible pair of words. This reformulation of the task outperforms all previous approaches over five different language pairs with varying levels of supervision.

% This reformulation overcomes the limitations of existing methods relying on softmax and span prediction allowing us to outperforms previous works in problems of various degree of supervision on five different language pairs.

\section{Related Work}

% \begin{itemize}
%     \item Transformer based word alignment methods outperformed traditional statistical methods such as giza++ and fastalign.
%     \item Multiple methods to use mPLMs contextualized embeddings (simalign, awesome, labse)
%     \item AwesomeAlign => parallel data
%     \item LaBSE => sentence level objective
%     \item Approaches do not require supervised data to perform well
% \end{itemize}

Recently, mPLM based approaches have significantly outperformed bilingual statistical methods \cite{och-ney-2003-systematic, dyer2013simple, ostling2016efficient} and bilingual neural methods \cite{garg-etal-2019-jointly, zenkel-etal-2020-end, chen-etal-2020-accurate, chen-etal-2021-mask, zhang2021bidirectional}. Among those approaches, we distinguish methods that achieve good performance without relying on manually annotated word alignment datasets \cite{jalili-sabet-etal-2020-simalign, dou-neubig-2021-word, wang-etal-2022-multilingual} from supervised methods that leverage existing word alignment datasets to train high performing word aligners \cite{nagata-etal-2020-supervised,wu-etal-2023-wspalign}. 

The first type of method relies on the approach of SimAlign \cite{jalili-sabet-etal-2020-simalign} which proposes to induce word alignment from the contextualized word embeddings of mPLMs pre-trained on non-parallel data. AwesomeAlign \cite{dou-neubig-2021-word} builds on top of this approach and proposes to fine-tune mPLMs on parallel text with different objectives to improve the quality of the contextualized word embeddings. More recently, AccAlign \cite{wang-etal-2022-multilingual} showed that models trained to learn language-agnostic sentence-level embeddings also learn strong language-agnostic word-level embeddings and set the state of the art in the zero-shot setting. Also, \citet{wang-etal-2022-multilingual} show that fine-tuning on existing word alignment datasets improves performance of AccAlign on language pairs unseen during word alignment fine-tuning. Our method is different from this body of work because our training and inference objective differ. We formalize word alignment as a binary sequence labeling task while we can see those methods as framing word alignment as a token retrieval task. 

In terms of approaches trained and evaluated in a supervised setting SpanAlign \cite{nagata-etal-2020-supervised} formalizes word alignment as a collection of SQuAD-style span prediction problems which differs from our binary classification objective. However, this method falls short in zero-shot and few-shot when word alignment training data is scarce. To remedy this problem, WSPAlign \cite{wu-etal-2023-wspalign} introduces a pre-training method based on weak supervision that significantly improves performance for all amounts of training data. 

% Our pre-training method differs from the one used in WSPAlign as we leverage of existing word alignment datasets in different languages to pre-train our model.  

% \bfs{Modern word alignment methods follow the ubiquitious trend of pre-trained transformers and fall into two primary categories.  The first uses pairs of parallel sentences in a dual-encoder (cite) model framework, relying on the assumption that the similarity learned at the sentence level is echoed at a token level. \cite{jalili-sabet-etal-2020-simalign, dou-neubig-2021-word, wang-etal-2022-multilingual}}. 

\begin{table*}
\small
\centering
\begin{tabular}{lllllll}
    \toprule
    {} & de-en & ro-en & fr-en & zh-en & ja-en & avg\\
    \midrule
    \multicolumn{1}{l}{Bilingual Methods} & \multicolumn{6}{c}{AER(\%)} \\
    \midrule
    \textsc{fast-align \cite{dyer2013simple}} & 27.0 & 32.1 & 10.5 & 38.1 & 51.1 & 31.8 \\
    \textsc{GIZA++ \cite{och-ney-2003-systematic}} & 20.6 & 26.4 & 5.9 & 35.1 & 48.0 & 27.2 \\
    \textsc{eflomal \cite{ostling2016efficient}} & 22.6 & 25.1 & 8.2 & 28.7 & 47.5 & 26.4 \\
    \textsc{MASK-ALIGN \cite{chen-etal-2021-mask}} & 14.4 & 19.5 & 4.4 & - & - & - \\
    \textsc{BTBA \cite{zhang2021bidirectional}} & 14.3 & 18.5 & 6.7 & - & - & - \\
    \midrule
    \multicolumn{1}{l}{Multilingual Methods}  & \multicolumn{6}{c}{AER(\%)} \\
    \midrule
    \textsc{SimAlign \cite{jalili-sabet-etal-2020-simalign}} & 18.8 & 27.2 & 7.6 & 21.6 & 46.6 & 24.4 \\
    \textsc{WSP \cite{wu-etal-2023-wspalign}} & 16.4 & 20.7 & 9.0 & 21.6 & 43.0 & 22.1 \\
    \textsc{AwesomeAlign \cite{dou-neubig-2021-word}} & 15.2 & 25.5 & 4.0 & 13.4 & 40.6 & 19.7 \\
    \textsc{SpanAlign-Align6} & 13.3 & 25.9 & 2.9 & 15.5 & 41.0 & 19.7 \\
    \textsc{AccAlign \cite{wang-etal-2022-multilingual}} & 13.5 & 20.8 & 2.8 & 11.3 & 37.0 & 17.1 \\
    \textsc{BinaryAlign} & \textbf{11.6} & \textbf{19.1} & \textbf{1.5} & \textbf{9.0} & \textbf{29.2} &  \textbf{14.1}\textbf{($\downarrow$3.0)} \\
    \bottomrule
\end{tabular}
 \caption{Comparison of AER(\%) between our method (BinaryAlign) and previous works on five unseen word alignment language pairs (zero-shot cross-lingual transfer). We highlight the best performance for each language pair in bold font. The arrow shows the performance improvement when compared to previous state-of-the-art.}
\label{table:zero_shot}
\end{table*}

\section{Method}\label{method}

Given a sentence $\mathcal{X}$ with $n$ words and a translation into another language $\mathcal{Y}$ with $m$ words, the task of word alignment is to produce an $n$ by $m$ adjacency matrix for the bipartite graph with the words of $\mathcal{X}$ on one side and the words of $\mathcal{Y}$ on the other (refer to Figure \ref{fig:example} for an illustration).  As our model will employ commonly used subword tokenization preprocessing we assume access to an invertible tokenizer, often implemented in practice with a list of subword units, a greedy algorithm for subword chunking, and leading symbols to denote word continuation in the vocab file \cite{sennrich-etal-2016-neural, wu2016google}.

We present BinaryAlign, a novel word alignment approach using a binary sequence labeling model, shown in Figure \ref{fig:model}.  The inputs to this model are a subword tokenized source sentence $X=x_1,x_2,...,x_{|X|}$, a subword tokenized target sentence $Y=y_1,y_2,...,y_{|Y|}$, and a reference word $w_X = x[i:j]$ which is a subspan of X.  We model the distribution of a binary alignment vector $A$ of size $|Y|$ in which each entry $a_k$ indicates if the word in $Y$ that contains $y_k$ is aligned to $w_X$.

We first preprocess $X$ by surrounding $w_X$ with unique separator tokens and then cross-encode the source and target sentences with an mPLM.  For each token $y_k$ in the target sentence we pass its final encoded representation through a linear layer to produce a single logit $z_k$.  We model $a_k$ with a logistic function using $z_k$ as its parameter:
\begin{equation}
    p(a_k = 1 | w_X) = \frac{1} {1 + e^{-z_k}}
\end{equation}
A supervised signal for token level alignments $a_k$ is easily divined from word alignment data and so this form is sufficient to estimate the parameters of the model.  However, our true inference-time goal of word to word alignment requires the use of additional heuristics.  To motivate these heuristics we formalize $\mathcal{W}$ as the inverse of the surjective mapping between the token indices in $Y$ and its corresponding words; for any subspan $w_Y$ of $Y$, $\mathcal{W}(w_Y)$ returns the token indices in $Y$ that compose $w_Y$.  

% We define the probability of the event $a^{\prime}_{ij}$ that there exists an alignment between word $w_i$ in $X$ and word $w_j$ in $Y$ as 
% \begin{equation}
%     p(a^{\prime}_{ij}) = \max_{\forall k \in \mathcal{W}(w_j)}p(a_k = 1 | w_i)
% \end{equation}

Given an aggregation function $agg$, we define the probability of the event $a^{\prime}$ that there exists an alignment between word $w_X$ in $X$ and word $w_Y$ in $Y$ as 
\begin{equation}
    p(a^{\prime}) = \underset{\forall k \in \mathcal{W}(w_Y)}{agg}p(a_k = 1 | w_X)
\end{equation}
Preliminary experiments suggest that the maximum aggregation strategy yields slightly superior performance compared to the mean and minimum aggregation strategies; hence, we adopt it for all subsequent experiments.

Word alignment in its general form is a symmetric problem in that we would expect the same answer if the source and target were swapped.  However, like most leading word alignment methods, our method is asymmetric; the source and target sentences are handled differently.  This deficiency is empirically detrimental to performance with the common remedy being to perform alignment in both directions and then to merge the two predictions in some manner.

We use the following symmetrization technique: letting $p_{X \rightarrow Y}$ denote the use of $X$ as the source and $Y$ as the target sentence, we average $p_{X \rightarrow Y}(a^{\prime})$ and $p_{Y \rightarrow X}(a^{\prime})$ and apply a threshold decision rule to make our final inference prediction.  While outside the scope of this study, we note that various other options exist and have been explored in previous work such as bidirectional average \cite{nagata-etal-2020-supervised}  or intersection, union and grow-diag-final: the default symmetrization heuristics supported in Moses \cite{koehn2007moses}.

% that threshold $p_{X \rightarrow Y}(a^{\prime}_{ij})$ and $p_{Y \rightarrow X}(a^{\prime}_{ji})$ and then average the two probabilities

\vspace{3em}

\section{Experiments}

\subsection{Datasets}\label{eval_details}

% \bfs{suggestion - in datasets, just describe the datasets.  Then, do a section called "experimental setups" right after that explains the two setups, and also add a header to each setup saying what the big picture is - i.e. for align6 we train on supervized data but eval on other languages, and for the other setup its good old fashioned supervized learning...but there's the matter of do you also train on other languages or not (Bilingual vs Multilingual)}

We use seven datasets of manually annotated word alignment data for our main experiments: French-English (fr-en), Chinese-English (zh-en), Romanian-English (ro-en), Japanese-English (ja-en), German-English (de-en), Swedish-English (sv-en) and ALIGN6. 
The ja-en data comes from the KFTT word alignment data \cite{neubig11kftt}, while the ro-en and fr-en data are taken from \citet{mihalcea2003evaluation} and the de-en data is provided by \citet{vilar2006aer}. Also, the zh-en data is obtained from the TsinghuaAligner website\footnote{\url{http://nlp.csai.tsinghua.edu.cn/~ly/systems/TsinghuaAligner/TsinghuaAligner.html}} and the sv-en dataset\footnote{\url{https://www.ida.liu.se/divisions/hcs/nlplab/resources/ges/}} from \citet{holmqvist2011gold}. Finally, ALIGN6 \cite{wang-etal-2022-multilingual} is the combination of six different word alignment datasets featuring Dutch-English \cite{macken2010annotation}, Czech-English \cite{marevcek2011automatic}, Hindi-English \cite{aswani2005aligning}, Turkish-English \cite{cakmak2012word}, Spanish-English \cite{graca2008building} and Portuguese-English \cite{graca2008building}. 

In addition, we use the Finnish-Greek (fi-el) and the Finnish-Hebrew (fi-he) word alignment test dataset from \citet{imani2021graph} to experiment on non-English language pairs. Note that all datasets are the same ones used in \citet{wang-etal-2022-multilingual}.

% \ma{\noindent\textbf{Few-shot and supervised setting: } We follow the protocol of \citet{wu-etal-2023-wspalign} for our few-shot and supervised experiments. For de-en, ro-en, ja-en and fr-en. For ja-en, we train on all eight dev set files, we use four test set files for testing and the remainder three test files for validation. We separate the de-en, ro-en and fr-en data into a training and test sets.  We fine-tune using 300 sentences for fr-en and de-en, while we use 150 sentences ro-en. All remaining sentences are used for testing. We made the same splits\footnote{https://huggingface.co/datasets/qiyuw/wspalign\_ft\_data} as \citet{wu-etal-2023-wspalign}. For the zh-en data, we use the 450 sentences of the dev set provided in v1 of the TsinghuaAligner website.}

\subsection{Experimental setup}\label{experimental setup}

\noindent\textbf{Unseen alignment experiments: } In our unseen alignment experiments, models are not fine-tuned on manual word alignment data of the tested language pair. This replicates a common real-world situation in which alignment data set is not available for a language pair and models must leverage knowledge gleaned from other language pairs and pre-training. This setting is usually referred as zero-shot cross-lingual transfer \cite{conneau-etal-2020-unsupervised, chi-etal-2021-infoxlm}. We follow \citet{wang-etal-2022-multilingual} and fine-tune our model on ALIGN6 and use sv-en as our validation set. We evaluate our method following the experimental protocol of previous work \cite{dou-neubig-2021-word, wang-etal-2022-multilingual} and use de-en, ro-en, fr-en, zh-en and ja-en for testing. Note that those language pairs are not included in ALIGN6.

\noindent\textbf{Few-shot and fully supervised experiments: } We follow the protocol of \citet{wu-etal-2023-wspalign} for our few-shot and fully supervised experiments on de-en, ro-en, fr-en and ja-en. For ja-en, we train on all eight dev set files, we use four test set files for testing, and the remaining three test files for validation. We separate the de-en, ro-en and fr-en data into training and test sets.  We fine-tune using 300 sentences for fr-en and de-en, while we use 150 sentences for ro-en. All remaining sentences are used for testing. Note that we made the same splits\footnote{https://huggingface.co/datasets/qiyuw/wspalign\_ft\_data} as \citet{wu-etal-2023-wspalign}. For the zh-en data, we leverage the datasets provided in v1 of the TsinghuaAligner website. We use their dev set and test set as our training and test set respectively. Both contain 450 sentences. 

In our non-English experiments, we evaluate on Finnish to Greek (fi-el) and Finnish to Hebrew (fi-he) data. For fi-el we use 400 samples for training and test on the 391 remaining samples, and for fi-he, we have 1780 samples for training and 450 for test. We use 32 training samples for all our few-shot experiments.

A detailed account of the number of sentence pairs for each dataset and settings used in our experiments is available in appendix.

\begin{table*}
\small
\centering
\begin{tabular}{lllllll}
    \toprule
     & de-en & ro-en & fr-en & zh-en & ja-en & avg \\
    \midrule
    \midrule
    \multicolumn{1}{l}{Few-Shot Supervision} & \multicolumn{6}{c}{AER(\%)} \\
    \midrule
    \textsc{AccAlign \cite{wang-etal-2022-multilingual}} & 11.9 & 16.5 & 2.7 & 10.7 & 35.3 & 15.4  \\
    \textsc{SpanAlign \cite{nagata-etal-2020-supervised}} & 15.4 & 14.8 & 8.0 & 16.0 & 43.3 & 19.5 \\
    \textsc{WSPAlign \cite{wu-etal-2023-wspalign}} & 10.2 & 10.9 & 3.8 & 11.1 & 28.2 & 12.8 \\
    \textsc{BinaryAlign-noPre} & 9.6 & 10.1 & 5.1 & 8.6 & 25.3 &  11.7 \\
    \textsc{BinaryAlign} & \textbf{7.6} & \textbf{8.8} & \textbf{2.5} & \textbf{6.7} & \textbf{22.8} & \textbf{9.7}\textbf{($\downarrow$3.1)}  \\
    \midrule
    \midrule
    \multicolumn{1}{l}{Full Supervision} & \multicolumn{6}{c}{AER(\%)} \\
    \midrule
    \textsc{AccAlign \cite{wang-etal-2022-multilingual}} & 11.7 & 16.8 & 2.6 & 10.1 & 31.2 & 14.5   \\
    \textsc{SpanAlign \cite{nagata-etal-2020-supervised}} & 14.4 & 12.2 & 4.0 & 8.9 & 22.4 & 12.4  \\
    \textsc{WSPAlign \cite{wu-etal-2023-wspalign}} & 11.1 & 8.6 & 2.5 & 7.6 & 16.3 & 9.2  \\
    \textsc{BinaryAlign-noPre} & 8.0 & 7.8 & \textbf{1.7} & 5.2 & 14.2 & 7.4  \\
    \textsc{BinaryAlign} & \textbf{7.7} & \textbf{7.3} & 1.9 & \textbf{4.8} & \textbf{13.9} & \textbf{7.1}\textbf{($\downarrow$2.1)}  \\
    \bottomrule
\end{tabular}
 \caption{Comparison of AER(\%) between the proposed method (BinaryAlign) and previous works with few-shot and full supervision. We highlight in bold the best performance in each problem. The arrow shows the performance improvement over previous state-of-the-art.}
\label{table:supervised}
\end{table*}

\subsection{Baseline methods}

\noindent\textbf{Unseen alignment experiments: } In this setting, we compare BinaryAlign to the three main bodies of research that evaluate on unseen word alignment language pairs. The first one corresponds to the historical bilingual statistical methods. We report GIZA++ \cite{och-ney-2003-systematic}, eflomal \cite{ostling2016efficient} and fast-align \cite{dyer2013simple} which are the best-known statistical methods. 

Bilingual neural methods represent the second body of work. For this, we report MASK-ALIGN \cite{chen-etal-2021-mask} and BTBA-FCBO-SST \cite{zhang2021bidirectional}, the two best performing bilingual neural methods. 

Finally, we compare to three methods relying on contextualized word embeddings of mPLMs: SimAlign \cite{jalili-sabet-etal-2020-simalign}, AwesomeAlign \cite{dou-neubig-2021-word} and AccAlign \cite{wang-etal-2022-multilingual}. Note that AccAlign leverages ALIGN6 and is the state-of-the-art method for unseen alignment.

We reimplemented AccAlign using their adapter method on ALIGN6 as done in there paper. For other baselines, we quote the results from \citet{dou-neubig-2021-word} for bilingual statistical methods, from \citet{wang-etal-2022-multilingual} for AwesomeAlign, SimAlign, and the bilingual neural methods. We also report the zero-shot performance of WSPAlign \cite{wu-etal-2023-wspalign} that we computed using the checkpoints provided by the authors. Also, we train SpanAlign on ALIGN6, the same dataset that is used to train AccAlign and BinaryAlign in this setup. Comparing to this baseline allows us to determine if our formalization of word alignment can better leverage existing word alignment datasets than SQuAD-style span prediction techniques. 

% However, we acknowledge that WSPAlign is not made to be used in zero-shot and report the results for information purposes.

\noindent\textbf{Few-shot and fully supervised experiments: } We compare to SpanAlign \cite{nagata-etal-2020-supervised} and WSPAlign \cite{wu-etal-2023-wspalign}, the two state of the art supervised word alignment techniques. Also, we further fine-tune AccAlign on our language specific training sets. Comparing to this baseline is important as it allows us to determine if our proposed method performs better than state of the art contextualized word embedding extraction techniques when we have access to manual word alignments of the same language pair. Also, SpanAlign and AccAlign have not been compared in previous studies.

% \begin{itemize}
%     \item AccAlign \cite{wang-etal-2022-multilingual}: We further fine-tune AccAlign on our language specific training sets. Comparing to this strong baseline allows us to determine if our proposed method performs better than state of the art contextualized word embedding extraction techniques when we have access to manual word alignments of the same language.
%     \item SpanAlign \cite{nagata-etal-2020-supervised}: The authors propose to formalize word alignment as a SQuAD v2.0 style question answering task. They reached state of the art performance in supervised setting without any pre-training. Comparing this method to our technique enables us to ascertain the effectiveness of our suggested approach.
%     \item WSPAlign \cite{wu-etal-2023-wspalign}: This work proposes a method to improve the performance of SpanAlign by pre-training using a weakly supervised dataset. Comparing to this method not only allows to determine if our proposed method performs better than Span prediction, but also to assess if our pre-training on other language word alignment data performs better than training on weakly supervised datasets.
% \end{itemize}

We reimplemented WSPAlign and SpanAlign for all our few-shot experiments using the source code provided by WSPAlign authors\footnote{https://github.com/qiyuw/WSPAlign}. We quote the results from \citet{wu-etal-2023-wspalign} for supervised fine-tuning of SpanAlign and WSPAlign on de-en, ro-en, ja-en, fr-en and reimplement them for zh-en as we use a different train and test dataset than the original papers.

\subsection{Fine-tuning setups} 

We did not perform extensive hyper-parameter tuning of our methods. We arbitrarily use a learning rate of $2e^{-5}$, a batch size of $8$ for fine-tuning and pre-training, and a threshold of $0.5$ for inference. We train all our models for 5 epochs, except for few-shot learning without pre-training on ALIGN6 where we train for 25 epochs. Results would improve with hyper-parameter tuning on a large validation set. We use mdeberta-v3-base \cite{he2021debertav3} as our mPLM. We discuss the choice of mPLM in section  \ref{subsec:mPLM}. In supervised settings we report both BinaryAlign and BinaryAlign-noPre, our method with and without pre-training on ALIGN6. Note that pre-training our model on ALIGN6 is the default setting in BinaryAlign.

\begin{table}
\small
\centering
\begin{tabular}{llll}
    \toprule
      & & fi-el & fi-he \\
    \midrule
    \makecell{Unseen\\alignment} & \makecell{AccAlign\\WSPAlign\\BinaryAlign} & \makecell{37.0\\34.9\\\textbf{24.3}} & \makecell{73.7\\74.7\\\textbf{40.4}}\\
    \midrule
    few-shot & \makecell{AccAlign\\WSPAlign\\BinaryAlign} & \makecell{30.3\\14.8\\\textbf{10.8}} & \makecell{51.5\\27.0\\\textbf{16.9 }}\\
    \midrule
    full & \makecell{AccAlign\\WSPAlign\\BinaryAlign} & \makecell{25.1\\8.6\\\textbf{6.2}} & \makecell{40.15\\10.8\\\textbf{7.2}}\\
    \bottomrule
\end{tabular}
 \caption{Comparison of AER(\%) between our method (BinaryAlign) and previous works in different settings on two non-English language pairs.}
\label{table:non_English}
\end{table}

\subsection{Word Alignment Evaluation Metric}

Following previous works \cite{wang-etal-2022-multilingual, dou-neubig-2021-word}, we evaluate performance using Alignment Error Rate (AER). Given a set of sure alignments (S), possible alignments (P) and predicted alignments (H) we can calculate AER as follows:
\begin{equation*}
    AER(S,P,H) = 1-\frac{|H\cap S|+|H\cap P|}{|H|+|S|}.
\end{equation*}
Following the protocol in \citet{wu-etal-2023-wspalign} we use only sure alignments for training but we evaluate on both sure and possible alignments when the distinction is available.

%In our settings, ALIGN6, zh-en, fr-en and de-en make a distinction between sure and possible alignments. We follow \citet{wu-etal-2023-wspalign} and fine-tune our models on sure data only but evaluate on sure and possible alignments.

\begin{table*}
\small
\centering
\begin{tabular}{lllllll}
    \toprule
      & de-en & ro-en & fr-en & zh-en & ja-en & avg \\
    \midrule
    \midrule
   \multicolumn{1}{l}{Few-shot Supervision} & \multicolumn{6}{c}{AER(\%)} \\
    \midrule
    mBERT & 9.8 & 12.7 & 3.5 & 9.3 & 26.6 & 12.4 \\
    mDeBERTa & \textbf{7.6} & 8.8 & 2.5 & 6.7 & 22.8 & 9.7 \\
    LaBSE & 7.9 & 9.3 & 2.4 & 6.4 & 23.4 & 9.9 \\
    XLM-RoBERTa-base & 8.4 & 9.3 & \textbf{2.4} & 8.5 & 31.1 & 11.9 \\
    XLM-RoBERTa-large & 7.7 & \textbf{8.4} & 3.1 & \textbf{6.0} & \textbf{21.8} & \textbf{9.4} \\
    \midrule
    \midrule
    \multicolumn{1}{l}{Full Supervision} & \multicolumn{6}{c}{AER(\%)} \\
    \midrule
    mBERT & 9.4 & 10.3 & 3.1 & 6.6 & 16.8 & 9.2 \\
    mDeBERTa & \textbf{7.7} & 7.3 & \textbf{1.9} & 4.8 & 13.9 & 7.1 \\
    LaBSE & \textbf{7.7} & 7.3 & 2.4 & 5.1 & 14.7 & 7.4  \\
    XLM-RoBERTa-base & 8.4 & 7.4 & 2.3 & 5.9 & 16.7 & 8.1 \\
    XLM-RoBERTa-large & \textbf{7.7} & \textbf{6.9} & 2.2 & \textbf{4.4} & \textbf{13.6} & \textbf{7.0} \\
    \bottomrule
\end{tabular}
 \caption{Comparison of AER(\%) of BinaryAlign under few-shot and full supervision using different mPLMs.}
\label{table:mPLMs}
\end{table*}

\begin{table}
\small
\centering
\begin{tabular}{lccc}
    \toprule
    Test set & Direction & SpanAlign & BinaryAlign  \\
    \midrule
    de-en & \makecell{de-en\\en-de\\sym} & \makecell{83.6($\downarrow$2.0)\\84.5($\downarrow$1.1)\\85.6} & \makecell{91.9($\downarrow$0.4)\\92.2($\downarrow$0.1)\\92.3}\\
    \midrule
    ro-en & \makecell{ro-en\\en-de\\sym} & \makecell{85.5($\downarrow$2.3)\\86.7($\downarrow$1.1)\\87.8} & \makecell{92.6($\downarrow$0.1)\\92.2($\downarrow$0.5)\\92.7}\\
    \midrule
    fr-en & \makecell{fr-en\\en-fr\\sym} & \makecell{85.0($\downarrow$1.7)\\85.4($\downarrow$1.3)\\86.7} & \makecell{98.0($\downarrow$0.2)\\98.0($\downarrow$0.2)\\98.2}\\
    \midrule
    ja-en & \makecell{ja-en\\ja-en\\sym} & \makecell{80.2($\uparrow$2.4)\\65.2($\downarrow$12.4)\\77.6} & \makecell{85.6($\downarrow$0.5)\\85.6($\downarrow$0.5)\\86.1}\\
    \bottomrule
\end{tabular}
 \caption{Comparison of our method (BinaryAlign) and reported results for SpanAlign \cite{nagata-etal-2020-supervised} when using symmetrization. We report the F1 score for each direction and the best symmetrized result (sym) from all explored heuristics. See appendix for details on our results and metric.}
\label{table:symmetrization_short}
\end{table}

\subsection{Results and Discussion}

We performed several experiments to validate BinaryAlign. First we compare our method to other state-of-the-art methods in three different levels of supervision: full available supervision, few-shot (32 samples), and unseen languages (zero-shot cross-lingual transfer). We also evaluate on non-English language pairs. Next, we evaluate the impact of choices for mPLM foundation and symmetrization. Finally, we discuss how our problem formulation compares to span prediction and contextualized word embedding based approaches in different situations.

\subsubsection{Comparison to State-of-the-Art}

%As described in section \ref{eval_details}, we evaluate our method on three different settings: zero-shot where we use 0 examples in the train dataset, few-shot where we use 32 examples, and supervised where we use the full amount of examples available depending on the training dataset size.

\noindent\textbf{Unseen alignments: } As a first experiment, we apply all methods to new language pairs without performing word alignment fine-tuning on the tested language pair as explained in \ref{experimental setup}. Table \ref{table:zero_shot} reports the AER of all methods. BinaryAlign is the new state-of-the-art on all language pairs. In particular, it outperforms AccAlign by 3.0 points of AER on average. Since AccAlign and BinaryAlign share the same pre-training dataset (ALIGN6), this indicates that our word alignment problem formulation performs better than inducing word alignment from contextualized word embeddings. This is also true when comparing with SpanAlign pre-trained on the same data (ALIGN6). This indicates that our formalization of word alignment promotes learning more language-agnostic signals from word alignment datasets when compared to existing methods.

% As a first experiment, we apply word alignment methods to a new language pair without performing word alignment fine-tuning on the tested language pair as explained in \ref{experimental setup}. Table \ref{table:zero_shot} reports the AER of all methods. BinaryAlign is the new state-of-the-art on all language pairs. In particular, it outperforms AccAlign by 3.0 points of AER on average. Since AccAlign and BinaryAlign share the same pre-training dataset ALIGN6, this indicates that our word alignment problem formulation performs better than inducing word alignment from contextualized word embeddings. In addition BinaryAlign outperforms SpanAlign when pre-trained on ALIGN6 indicating that our formalization of word alignment tend to learn a more language-agnostic signal from existing word alignment datasets than previous supervised methods.
\medskip
\noindent\textbf{Full and few shot supervision:} We compare BinaryAlign to the other baseline methods after fine-tuning on alignment data with few samples and the whole training data set. Table \ref{table:supervised} shows the results for both supervision levels.

Our method achieves new state-of-the-art on all tested language pairs and with both levels of supervision. On average it outperforms WSPAlign, the previous state-of-the-art, by 2.1 points of AER with full supervision and 3.1 with few-shot supervision.

Even without pre-training on ALIGN6 our method outperforms all methods. Given that WSPAlign was pre-trained on 2 millions samples, it indicates that BinaryAlign promotes sample efficiency. 

Finally, we highlight that by using only 32 samples for few-shot supervision BinaryAlign outperforms SpanAlign regardless of pre-training. This reinforces the performance improvement of formalizing word alignment as a binary token classification objective over span prediction. 

% Table \ref{table:supervised} shows the comparison of our method against previous works in supervised settings.

% \noindent\textbf{Full supervision: } Our method achieves new state-of-the-art on all tested language pairs. Specifically, it outperforms WSPAlign by on average 2.1 points of AER. We note that even without pre-training on ALIGN6, our method outperforms SpanAlign and WSPAlign by 5 and 1.8 points of AER respectively. As WSPAlign was pre-trained on 2 millions samples, this clearly indicates the superiority and sample efficiency of BinaryAlign.

% \noindent\textbf{Few Shot: } As in the full supervision setting, BinaryAlign achieves new state-of-the-art on all tested language pairs. It is important to highlight that using only 32 samples, our method, with and without pre-training, performs on average better than SpanAlign in full supervision. This reinforces the performance improvement of formalizing word alignment as a binary token classification objective. 
\medskip
\noindent\textbf{Impact of pre-training on other languages:} Table \ref{table:supervised} reports the results of our method with and without pre-training on ALIGN6. We conclude that pre-training improves AER with few-shot as well as full supervision. However, we observe a smaller improvement with full supervision which suggests that the benefit of pre-training on other languages is inversely correlated with the amount of in-domain word alignment data. While pre-training encourages sample efficiency, we did not find any indication that it could hinder performance.

\medskip
\noindent\textbf{Non-English language pairs:} Because usually mPLMs perform better in English it is important to investigate how our method performs on non-English language pairs. Table \ref{table:non_English} reports results on language pairs excluding English. We used the checkpoint\footnote{https://huggingface.co/qiyuw/WSPAlign-xlm-base} provided by the authors of WSPAlign since paragraph pairs in Finnish-Greek and Hebrew are difficult to obtain for training. Our results show that BinaryAlign outperforms WSPAlign and AccAlign for all degree of supervision. Also, the AER in non-English language pairs seems to be similar to the AER of our main experiments on English-centric language pairs which shows that our method does not depend on English and is robust to variations in language family.
  
\subsubsection{Design Choices}\label{subsec:mPLM} 
\label{subsec:size_scaling} 
\begin{table*}
\small
\centering
\begin{tabular}{lllllll}
    \toprule
      & de-en & ro-en & fr-en & zh-en & ja-en & avg \\
    \midrule
    \midrule
    \multicolumn{1}{l}{\textbf{Untranslated words}} & \multicolumn{6}{c}{Correctly aligned words(\%)} \\
    \midrule
    Number of occurances & 2085 & 974 & 674 & 5882 & 6204 & 3164 \\
    \midrule
    AccAlign & 74.1 & 79.5 & 82.2 & 75.9 & 75.0 & 77.3 \\
    SpanAlign & 81.5 & 85.2 & 88.6 & 79.3 & 86.6 & 84.3 \\
    BinaryAlign & \textbf{84.4} & \textbf{88.3} & \textbf{94.1} & \textbf{83.5} & \textbf{89.0} & \textbf{87.9} \\
    \midrule
    \midrule
    \multicolumn{1}{l}{\textbf{One-to-multiple alignments}} & \multicolumn{6}{c}{Correctly aligned words(\%)} \\
    \midrule
    Number of occurances & 2079 & 1726 & 6159 & 1738 & 3937 & 3128 \\
    \midrule
    AccAlign & 13.8 & 5.3 & 1.2 & 44.9 & 11.8 & 15.4 \\
    SpanAlign & 27.6 & 11.0 & \textbf{5.6} & 48.8 & 21.6 & 22.9 \\
    BinaryAlign & \textbf{31.4} & \textbf{16.7} & 4.8 & \textbf{60.2} & \textbf{29.0} & \textbf{28.4} \\
    \midrule
    \midrule
    \multicolumn{1}{l}{\textbf{One-to-multiple non-contiguous words}} & \multicolumn{6}{c}{Correctly aligned words(\%)} \\
    \midrule
    Number of occurances & 383 & 410 & 565 & 179 & 405 & 388 \\
    \midrule
    AccAlign & 5.5 & 5.6 & 3.5 & 15.1 & 4.9 & 6.9 \\
    SpanAlign & 11.0 & 2.2 & 2.5 & 8.9 & 4.7 & 5.9 \\
    BinaryAlign & \textbf{21.7} & \textbf{5.1} & \textbf{7.1} & \textbf{26.3} & \textbf{7.4} & \textbf{13.5} \\
    \bottomrule
\end{tabular}
 \caption{Comparison of AccAlign, SpanAlign and BinaryAlign in complex word alignment situations. The three methods are pre-trained on ALIGN6 and evaluated on unseen alignments. See \ref{appendix:metrics} for details on our metric.}
\label{table:post_analysis}
\end{table*}

\noindent\textbf{mPLM Architecture:} Our proposed reformulation of the word alignment problem does not depend on a particular mPLM architecture. In this experiment, we investigate the impact of using different mPLMs. We explore five different mPLMs: XLM-RoBERTa (base and large)\cite{conneau-etal-2020-unsupervised}, LaBSE\footnote{https://huggingface.co/sentence-transformers/LaBSE} \cite{feng-etal-2022-language}, mDeBERTa-v3-base\footnote{https://huggingface.co/microsoft/mdeberta-v3-base} \cite{he2021debertav3} and mBERT\footnote{https://huggingface.co/bert-base-multilingual-cased} \cite{devlin-etal-2019-bert}.

Table \ref{table:mPLMs} reports AER of BinaryAlign using different mPLMs in few-shot and fully supervised settings. All these versions of BinaryAlign reach or surpass the previous state-of-the-art in terms of average AER on the five tested language pairs. This highlights that the improvement of our method over previous state-of-the-art is not explained by its reliance on a specific mPLM. 

While most mPLMs yield similar results, mBERT performs slightly worse than the others. This could be due to a poor parametrization given that we used the same hyper-parameter configuration for all mPLMs. This could also be explained by the training objective of the mPLMs or their capacity. For example we observe that scaling the size of XLM-RoBERTa has an effect on alignment performance. The base model has an approximately similar capacity as the other mPLMs and when we increase this capacity using the large model we obtain the best result over all mPLMs. We suspect that this effect could generalize to other mPLM architectures.

\medskip
\noindent\textbf{Symmetrization:} Here we investigate the impact of different symmetrization heuristics on our results. As stated in Section \ref{method} symmetrization consists in fusing the alignment obtained going from one language to another with the alignment obtained going in the inverse direction. We compare several symmetrization techniques: intersection, union, average (avg) and bidirectional average (bidi-avg) \cite{nagata-etal-2020-supervised}. 

In Table \ref{table:symmetrization_short} we compare results obtained from aligning in a single direction to the results obtained using the best symmetrization heuristics (full details available in the appendix). We report the F1 score (see \ref{appendix:metrics}) as done in \citet{nagata-etal-2020-supervised}. %We do not compare to AccAlign here because the technique encodes the source and target separately which means that the symmetrization technique has no impact on inference time. 

Our results indicate that for BinaryAlign, unidirectional alignment does not perform significantly worse (average of 0.3 points of F1 score) than symmetrized alignment. This is not the case for SpanAlign which gains 2.4 points of F1 score on average by applying symmetrization. Performing alignment in only one direction is interesting since it halves the inference time. 

% We investigate the impact of the choice of symmetrization heuristic. In particular, we report the F1 score on the two directions, using intersection, union, average (avg) and bidirectional average (bidi-avg) \cite{nagata-etal-2020-supervised}. Table \ref{table:symmetrization} shows that we can switch to a unidirectional setup with a negligible loss of on average 0.3 points of F1 score. The same can not be said about span based methods represented by SpanAlign where switching to a unidirectional setup leads to a loss of on average 2.3 points of F1 score. Having the possibility to switch to a unidirectional setup is interesting as it allows to decrease inference time by a factor of 2. 

\subsubsection{Post-analysis of errors}

In this section we analyze how the proposed problem formulation of BinaryAlign improves accuracy in complex word alignment situations. We inspect results in three situations: 1) words that are untranslated, also referred as null words \cite{jalili-sabet-etal-2020-simalign} (2) words that are aligned to multiple words (3) words that are aligned to multiple non contiguous words. For each situation, we report the percentage of correctly aligned words in Table \ref{table:post_analysis}. Details on how we computed our metric can be found in \ref{appendix:metrics}. Results indicate that our method handles these situations better than both competing methods. This is especially true when aligning multiple non contiguous words which was the main motivation for our reformulation. The prevalence of these situations in a given language pair modulates the performance gain of our method over the others.

\section{Conclusion}

We presented BinaryAlign, a novel word alignment training and inference procedure. In particular, we proposed to reformulate the word alignment problem as a binary token classification task. We showed that because of this reformulation BinaryAlign outperforms existing methods regardless of the degree of supervision. In addition we showed that it overcomes the inherent limitations of previous methods relying on span prediction and softmax. As a result, we made the word alignment task easier to tackle by using a single model for both high and low-resource languages. 

In the future we plan to explore the use of larger decoder-only or encoder-decoder models such as mT5 \cite{xue-etal-2021-mt5} to see how much alignment performance will increase. We also plan on investigating knowledge distillation techniques to improve the inference time of our method.

\section{Limitations}

%The main limitation of our proposed method relies on its inference cost. Our method requires $n+m$ forward passes which makes it slow for long sequences. However, we note that previous state-of-the-art supervised approaches have \ma{similar inference time}. Also, our work lacks experiments on extremely low-resource languages that the mPLM has not seen during pre-training \cite{ebrahimi-etal-2023-meeting}. Whether or not our method will rapidly adapt to those new languages is an open question \cite{garcia2021towards}. 

The inference cost is the main limitation of our method. When using symmetrization, it has to perform a forward pass for each word of both sentences, which can be slow with long sequences. However, this is a drawback that we share with previous state-of-the-art supervised approaches \cite{nagata-etal-2020-supervised,wu-etal-2023-wspalign}.

In addition, we did not experiment on extremely low-resource languages that the mPLM has not seen during pre-training \cite{ebrahimi-etal-2023-meeting}. While the benefits of our new formulation would likely apply to any language, it is unclear how our method will rapidly adapt the mPLM to new languages \cite{garcia2021towards}. 

In real-world applications, translations are often partial and noisy. Unfortunately, we could not evaluate the robustness of our method to different translation pair quality because this type of word alignment dataset does not exist.

\bibliography{anthology,custom}
\bibliographystyle{acl_natbib}

\clearpage

\appendix

\section{Appendix}
\label{sec:appendix}

\subsection{Experimental Environment}

For all our experiments we use one NVIDIA Quadro RTX 6000. Fine-tuning on ALIGN6 took 4 hours and 30 minutes while our fully supervised experiments took on average 20 minutes per dataset.

\subsection{Dataset statistics}

Table \ref{table:statistics} shows the number of samples that our training, validation and test set contains for all level of supervision. All dataset are the same as in \citet{wang-etal-2022-multilingual}. The de-en, ro-en, fr-en and ja-en train-test splits are the same as the one used in \citet{wu-etal-2023-wspalign}. We could not get the zh-en data used in \citet{wu-etal-2023-wspalign} because the dataset is not publicly available.

\subsection{Metric details }\label{appendix:metrics}

\subsubsection{F1 score}

Given a set of sure alignments (S), possible alignments (P) and predicted alignments (H), we can compute the Recall, Precision and F1 score as follows:

\begin{equation*}
    Recall(H, S) = \frac{|H\cap S|}{|S|}
\end{equation*} 
\begin{equation*}
    Precision(H, P) = \frac{|H\cap P|}{|H|}
\end{equation*} 
\begin{equation*}  
    F_1(H, S, P) = \frac{2*Precision*Recall}{Precision+Recall}
\end{equation*}

When $S==P$, we have:
\begin{equation*}
    AER(H, S, P) = 1-F_1(H,S,P)
\end{equation*}

\subsubsection{Post-analysis of errors}

\noindent\textbf{Untranslated words: } We report the number of untranslated words correctly aligned by the models over the total number of untranslated words. We consider a word to be correctly aligned if the model has not aligned it to any words in the corresponding translated sentence.

\noindent\textbf{One-to-multiple contiguous and non contiguous words: } In this case, we report the number of contiguous/non contiguous words correctly aligned by the model over the total number of contiguous/non contiguous words. We consider a word to be correctly aligned if the model has aligned it to the exact same set of ground truth aligned words.

\begin{table}
\small
\centering
\begin{tabular}{llllll}
    \toprule
    Test set & Method & SpanAlign & BinaryAlign  \\
    \midrule
    de-en & \makecell{De to En\\En to DE\\intersection\\union\\bidi-avg\\avg} & \makecell{83.6($\downarrow$2.0)\\84.5($\downarrow$1.1)\\84.0\\84.0\\85.6\\-} & \makecell{91.9($\downarrow$0.4)\\92.2($\downarrow$0.1)\\92.2\\91.9\\92.2\\92.3}\\
    \midrule
    ro-en & \makecell{Ro to En\\En to Ro\\intersection\\union\\bidi-avg\\avg} & \makecell{85.5($\downarrow$2.3)\\86.7($\downarrow$1.1)\\87.3\\85.0\\87.8\\-} & \makecell{92.6($\downarrow$0.1)\\92.2($\downarrow$0.5)\\92.2\\92.7\\92.7\\92.7}\\
    \midrule
    fr-en & \makecell{Fr to En\\En to Fr\\intersection\\union\\bidi-avg\\avg} & \makecell{85.0($\downarrow$1.7)\\85.4($\downarrow$1.3)\\86.7\\83.9\\86.2\\-} & \makecell{98.0($\downarrow$0.2)\\98.0($\downarrow$0.2)\\97.8\\98.2\\97.8\\98.0}\\
    \midrule
    ja-en & \makecell{Ja to En\\En to Ja\\intersection\\union\\bidi-avg\\avg} & \makecell{80.2($\uparrow$2.4)\\65.2($\downarrow$12.4)\\74.5\\71.1\\77.6\\ -} & \makecell{85.6($\downarrow$0.5)\\85.6($\downarrow$0.5)\\85.7\\85.5\\85.7\\86.1}\\
    \bottomrule
\end{tabular}
 \caption{F1 score comparison of our method (BinaryAlign) and SpanAlign using different symmetrization heuristics in supervised setting. For SpanAlign, we quote results from \citet{nagata-etal-2020-supervised}.}
\label{table:symmetrization}
\end{table}

\begin{table}
\small
\centering
\begin{tabular}{lllll}
    \toprule
      & Dataset & Train & Val & Test  \\
    \midrule
    \makecell{zero-shot\\cross-lingual transfer\\(unseen\\alignment)} & \makecell{Align6\\de-en\\ro-en\\fr-en\\zh-en\\ja-en\\sv-en} & \makecell{3,362\\-\\-\\-\\-\\-\\-} & \makecell{-\\-\\-\\-\\-\\-\\192} & \makecell{-\\508\\248\\447\\450\\582\\-}\\
    \midrule
    \midrule
    fully supervised & \makecell{de-en\\ro-en\\fr-en\\zh-en\\ja-en} & \makecell{300\\150\\300\\450\\653} & \makecell{-\\-\\-\\-\\225} & \makecell{208\\98\\147\\450\\357} \\
    \bottomrule
\end{tabular}
 \caption{Number of training, validation and test samples in different settings. We omit few-shot as it shares the same test set as the fully supervised setting but only use 32 samples for training in each language.}
\label{table:statistics}
\end{table}

\end{document}